\documentclass[runningheads]{llncs}

\PassOptionsToPackage{numbers,compress}{natbib}

\usepackage[utf8]{inputenc} 
\usepackage[T1]{fontenc}    
\usepackage{hyperref}       
\usepackage{url}            
\usepackage{booktabs}       
\usepackage{amsfonts}       
\usepackage{nicefrac}       
\usepackage{microtype}      

\usepackage[title]{appendix}
\usepackage{bbm}
\usepackage{amsmath,amsfonts,bm}
\usepackage{upgreek}
\usepackage{algorithm}
\usepackage{algorithmic}
\usepackage{microtype}
\usepackage{graphicx}
\usepackage{booktabs} 
\usepackage{xcolor}
\usepackage{setspace}
\makeatletter
\newcommand{\printfnsymbol}[1]{%
  \textsuperscript{\@fnsymbol{#1}}%
}
\makeatother

\title{Interpretable Neuron Structuring with Graph Spectral Regularization}
\author{
Alexander Tong\inst{1,}\thanks{\hspace{-3pt},\hspace{1pt}$^\dagger$ equal contribution} \and
David van Dijk\inst{2,}\printfnsymbol{1} \and
Jay S. Stanley III\inst{2} \and
Matthew Amodio\inst{1} \and
Kristina Yim\inst{2} \and
Rebecca Muhle\inst{2} \and
James Noonan\inst{2} \and
Guy Wolf\inst{3,}$^\dagger$,
Smita Krishnaswamy\inst{1,2,4,}$^\dagger$
}
\authorrunning{A. Tong et al.}

\institute{Yale Department of Computer Science \and
Yale Department of Genetics \and
Universit\'{e} de Montr\'{e}al, Dept. of Math. \& Stat. ; Mila \and
Corresponding Author \email{smita.krishnaswamy@yale.edu}
}

\begin{document}
\maketitle 
\begin{abstract}
While neural networks are powerful approximators used to classify or embed data into lower dimensional spaces, they are often regarded as black boxes with uninterpretable features. Here we propose \textit{Graph Spectral Regularization} for making hidden layers more interpretable without significantly impacting performance on the primary task. Taking inspiration from spatial organization and localization of neuron activations in biological networks, we use a graph Laplacian penalty to structure the activations within a layer. This penalty encourages activations to be smooth either on a predetermined graph or on a feature-space graph learned from the data via co-activations of a hidden layer of the neural network. We show numerous uses for this additional structure including cluster indication and visualization in biological and image data sets.

\end{abstract}

\section{Introduction}
\label{Introduction}
Common intuitions and motivating explanations for the success of deep learning approaches rely on analogies between artificial and biological neural networks, and the mechanism they use for processing information. However, one aspect that is overlooked is the spatial organization of neurons in the brain. Indeed, the hierarchical spatial organization of neurons, determined via fMRI and other technologies~\cite{ogawa_magnetic_1990,logothetis_neurophysiological_2001}, is often leveraged in neuroscience works to explore, understand, and interpret various neural processing mechanisms and high-level brain functions. In artificial neural networks (ANN), on the other hand, hidden layers offer no organization that can be regarded as equivalent to the biological one. This lack of organization poses great difficulties in exploring and interpreting the internal data representations provided by hidden layers of ANNs and the information encoded by them. This challenge, in turn, gives rise to the common treatment of ANNs as black boxes whose operation and data processing mechanisms cannot be easily understood. To address this issue, we focus on the problem of modifying ANNs to learn more interpretable feature spaces without degrading their primary task performance. 

While most neural networks are treated as black boxes, we note that there are methods in ANN literature for understanding the activations of filters in convolutional neural networks (CNNs)~\cite{lecun_backpropogation_1989}, either by examining trained networks~\cite{zeiler_visualizing_2013}, or by learning a better representation~\cite{liao_learning_2016,zhang_interpretable_2018,ross_right_2017,stone_teaching_2017,sabour_dynamic_2017}, but such methods rarely apply to other types of networks, in particular dense neural networks (DNNs) where a single activation is often not interpretable on its own. Furthermore, convolutions only apply to datatypes where we know the feature structure apriori, as in the case of images and natural language.  In layers of a DNN, there is no enforced structure between neurons. The correspondence between neurons and concepts is only determined based on the random initialization of the network. In this work, we encourage {\em structure between neurons} in the same layer, creating more localized and interpretable layers in dense architectures.    

More specifically we propose a \textit{Graph Spectral Regularization} to encourage arbitrary graph structure between neurons within a layer. The internal layers of a neural network are constrained to take the structure of a graph, with graph neighbors activating on similar inputs.  This allows us to map the activations of a given layer over the graph and interpret new input by examining the activations. We show that graph-structuring a hidden layer causes useful, interpretable features to emerge. For instance, we show that grid-structuring a layer of a classification network creates a structure over which convolution can be applied, and local receptive fields can be traced to understand classification decisions. 


While a majority of the time imposing a known graph structure gives interpretable results, there are circumstances where we would like to learn the graph structure from data. In such cases we can learn and emphasize the natural graph structure of the feature space. We do this by an iterative process of encoding the data, and modifying the graph based on the feature co-activation patterns. This procedure reinforces existing patterns in the data. This allows us to learn an abstracted graph structure of features in high-dimensional domains such as single-cell RNA sequencing. 

The main contributions of this work are as follows: 
(1) Demonstration of hierarchical, spatial, and smoothed feature maps for interpretability in dense networks.
(2) A novel method for learning and reinforcing the natural graph structure for complex feature spaces. 
(3) Demonstration of graph learning and abstraction on single-cell RNA-sequencing data.



\section{Related Work}
\label{sec:related_work}


\paragraph{Disentangled Representation Learning:} 
While there is no precise definition of what makes for a disentangled representation, the aim is to learn a representation that axis aligns with the generative factors of the data~\cite{higgins_towards_2018,achille_emergence_2017-1}. \cite{higgins_-vae_2017} suggest a way to disentangle the representation of variational autoencoders~\cite{kingma_auto-encoding_2013} with $\beta$-VAE. Subsequent work has generalized this to discrete representations~\cite{dupont_learning_2018}, and simple hierarchical representations~\cite{esmaeili_structured_2019}. These works focus on learning a single vector representation of the data, where each element represents a single concept. In contrast, our work learns a representation where groups of neurons may be involved in representing a single concept. Moreover, disentangled representation learning can only be applied to unsupervised models and only the most compressed level of either an autoencoder~\cite{higgins_-vae_2017} or generative adversarial network as in~\cite{chen_infogan_2016}, whereas graph spectral regularization (GSR) can be applied to any or all layers of the network.

\paragraph{Graph Structure in ANNs:} Graph based penalties have been used in the graph signal processing literature~\cite{belkin2004regularization,zhou2004regularization,shuman2013emerging}, but are rarely used in an ANN setting. In the biological data setting, \cite{min_network-regularized_2018-1} used a graph penalty in sparse logistic regression on gene expression data. Another way of utilizing graph structure is through graph convolutional networks (GCN). GCNs are a related body of work introduced by~\cite{gori_new_2005}, and expanded on by~\cite{scarselli_graph_2009}, but focus on a different set of problems (For an overview see~\cite{wu_comprehensive_2019}). GCNs require a known graph structure. We focus on learning a graph representation of general data. This learned graph representation could be used as the input to a GCN similar to our MNIST example.

\section{Enforcing Graph Structure}

We consider the intra-layer relationships between neurons or larger structures such as capsules. For a given layer of neurons we construct a graph $G=(V,E)$ with $V = \{v_1, \ldots, v_N\}$ the set of vertices and $E \subseteq V \times V$ the set of edges. Let $W$ be the weighted symmetric adjacency matrix of size $N \times N$ with $W_{ij} = W_{ji} \ge 0$ representing the weight of the edge between $v_i$ and $v_j$. The graph Laplacian $L$ is then defined as $L = D - W$ where $D_{ii} = \sum_{j} W_{ij}$ and $D_{ij} =0$ for $i \neq j$.

To enforce smoothing we use the Laplacian smoothing loss. On some activation vector $z$ and fixed Laplacian $L$ we formulate the graph spectral regularization function $G$ as:
\begin{equation}
    \label{eq:gsr}
    G(z,\mathbf{L}) = z^T \mathbf{L} z = \sum_{ij} W_{ij} ||z_i - z_j||
\end{equation}

Where $||\cdot||$ denotes the Frobenius norm. We add it to the reconstruction or classification loss with a weighting term $\alpha$. This adds an additional objective that activations should be smooth along the graph defined by $L$. This optimization procedure applies to any multi-layer model and valid graph Laplacian. We apply this algorithm to grid, and hierarchical graph structures on both autoencoder and classification dense architectures.

\begin{algorithm}[htb]
\caption{Graph Learning}
\label{alg:graph-learning}
\begin{algorithmic}
    \STATE{\bfseries Input} batches $x_i$, model $M$ with latent layer activations $z_i$, regularization weight $\alpha$.
    \STATE{\bfseries Pre-train} $M$ on $x_i$ with $\alpha = 0$
    \FOR{$i=1$ {\bfseries to} $T$}
        \STATE Create Graph Laplacian $L_i$ from activations $z_i$
        \FOR{$j=1$ {\bfseries to} $m$}
            \STATE Train $M$ on $x_i$ with $\alpha = w$ and $L = L_i$ with MSE + loss in eq. \ref{eq:gsr}
        \ENDFOR
    \ENDFOR
\end{algorithmic}
\end{algorithm}

\subsection{Learning and Reinforcing an Abstracted Feature-space Graph }~\label{sec:alg:learning}

Instead of enforcing smoothness over a fixed graph, we can learn a feature graph from the data (See Algorithm~\ref{alg:graph-learning}) using neural network activations themselves to bootstrap the process. Note, that most graph and kernel-based methods are applied over the space of observations but not over the space of features. One of the reasons is because it is even more difficult to define a distance between features than it is between observations. To circumvent this problem, we propose to learn a feature graph in the latent space of a neural network using feature co-activations as a measure of similarity. 

We proceed by creating a graph using feature activation similarity, then applying this graph using Laplacian smoothing for a number of iterations. This converges to a graph of a latent feature space at the level of granularity of the number of dimensions in the corresponding layer.

Our algorithm for learning the graph consists of two phases. First, a pretraining phase where the model is learned with no graph regularization. Second, we alternate between constructing the graph from the similarities of the embedding layer features and further training the network for reconstruction and smoothness on the graph. There are many ways to create a graph from the feature $\times$ datapoint activation matrix. We use an adaptive Gaussian kernel,
$$K(z_i, z_j) = \frac{1}{2} exp \bigg(- \frac{||z_i - z_j||^2_2}{\sigma_i^2} \bigg) + \frac{1}{2} exp \bigg(- \frac{||z_i - z_j||^2_2}{\sigma_j^2}\bigg)$$
where $\sigma_i$ is the adaptive bandwidth for node $i$ which we set as the distance to the $k^{th}$ nearest neighbor of feature. An adaptive bandwidth Gaussian kernel is necessary for general architectures as the scale of the activations is not fixed. Batch normalization can also be used to limit the activation scale. 

Since we are smoothing on the graph then constructing a new graph from the smoothed signal the learned graph converges to a steady state where the mean squared error acts as a repulsive force to stop the graph collapsing any further. We present the results of graph learning a biological dataset and show that the learned structure adds interpretability to the activations.
\section{Experiments}
\label{sec:results}
Through examples, we show that visualizing the activations of data on the regularized layer highlights relationships in the data that are not easily visible without it. We establish this with two examples on fixed graphs, then move to graphs learned from the structure of the data with two examples of hierarchical structure and two with progression structure.

\subsection{Fixed Structure}
Enforcing fixed graph structure localizes activations for similar datapoints to a region of the graph. Here we show that enforcing a 8x8 grid graph on a layer of a dense MNIST classifier causes receptive fields to form, where each digit occupies a localized group of neurons on the grid. This can, in principle, be applied to any neural network layer to group neurons activating to similar features. Like in FMRI data or a convolutional neural network, we can examine the activation patterns for each localized group of neurons. For a second example, we show the usefulness in encouraging localized structure on a capsulenet architecture~\cite{sabour_dynamic_2017}. Where we are able to create globally consistent structure for better alignment of features between capsules.

\begin{figure*}[ht]
    \begin{center}
    \includegraphics[width = 1\linewidth]{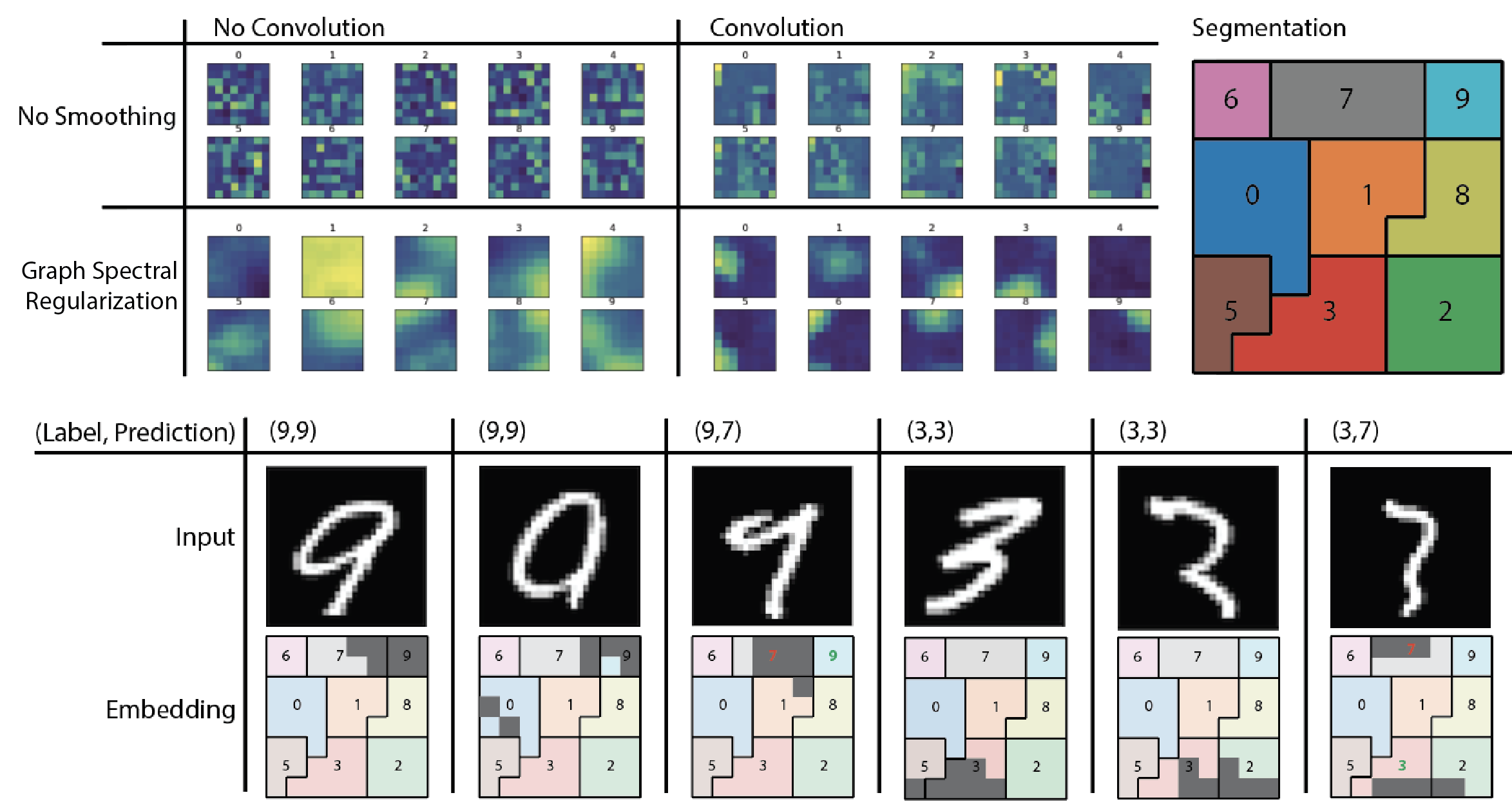}
    \end{center}
    \caption{Shows average activation by digit over an (8x8) 2D grid using graph spectral regularization and convolutions following the regularization layer. Next, we segment the embedding space by class to localize portions of the embedding associated with each class. Notice that the digit 4 here serves as the null case and does not show up in the segmentation. Finally, we show the top 10\% activation on the embedding of some sample images. For two digits (9 and 3) we show a normal input, a correctly classified but transitional input, and a misclassified input. The highlighted regions of the embedding space correlate with the semantic description of the input.}
    \label{fig:mnist_class_fig}
\end{figure*}
\subsubsection{Enforcing Grid Structure on Mnist.} Without GSR, activations are unstructured and as a result are difficult to interpret, in that it is difficult to visually identify even which class a digit comes from based on the activation pattern (See Fig.~\ref{fig:mnist_class_fig}). With GSR we can organize the activations making this representation more visually distinguishable. Since we can now take this embedding as an image, it is possible to use a standard convolutional architecture in subsequent layers in order to further filter the encodings. When we add 3 layers of 3x3 2D convolutions with 2x2 max pooling we see that representations for each digit are compressed into specific areas of the image. This leads to the formation of receptive fields over the network pertaining to similar datapoints. Using these receptive fields, we can now extract the features responsible for digit classification. For example, features that contribute to the activation of the top right of our grid we can associate with those features that contribute to being the digit 9. 

The activation patterns on the embedding layer correspond well to a human perception of the digit type. The 9 that is misclassified as 7 both has significant activation in the 7 region of the embedding layer, and looks visually close to a 7. We can now interpret the embedding layer as a sort of brain map, where the map can map regions of activations, to types of inputs. This is not possible in a standard neural network, where activations are not spatially organized.

\begin{figure*}[ht]
    \begin{center}
    \includegraphics[width=0.8 \linewidth]{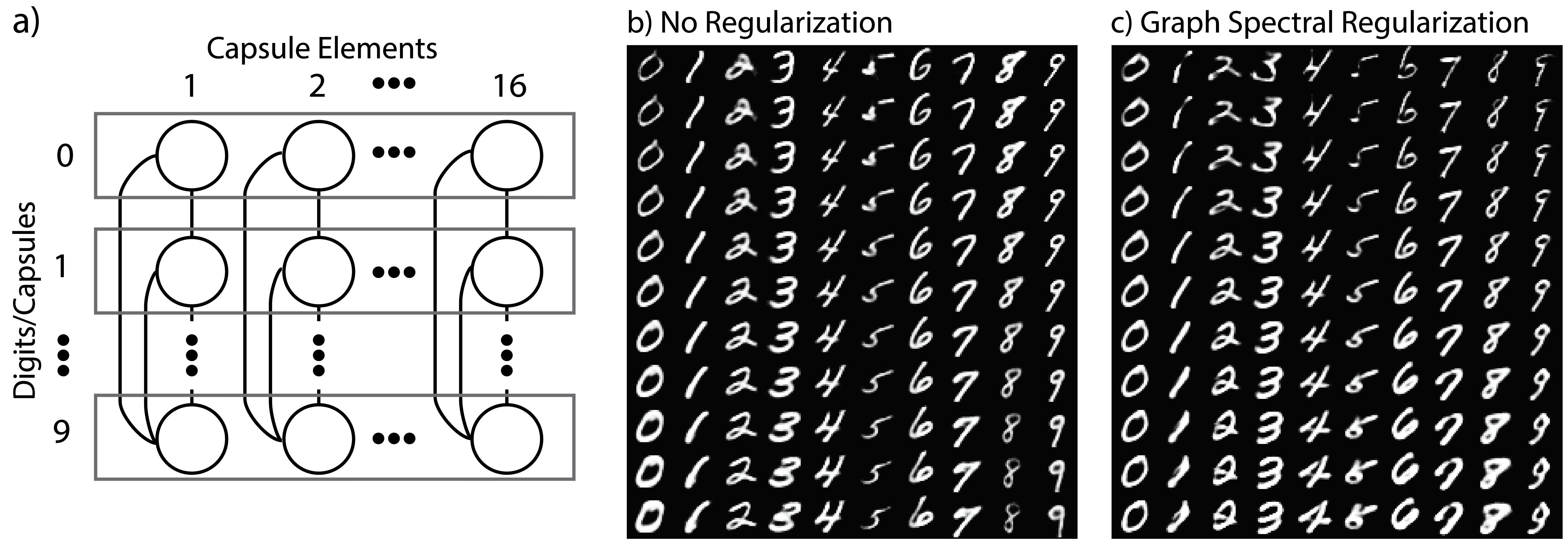}
    \end{center}
    \caption{(a) shows the regularization structure between capsules. (b-c) Show reconstruction when one of the 16 dimensions in the DigitCaps representation is tweaked by $0.05 \in [-0.25, 0.25]$. (b) Without GSR each digit responds differently to perturbation of the same dimension. With GSR (c) a single dimension represents line thickness across all digits. }
    \label{fig:caps_fig}
\end{figure*}
\subsubsection{Enforcing Node Consistency on Capsule Networks.}
\label{sec:results:learning}
Capsule networks~\cite{sabour_dynamic_2017} represent the input as a set of vectors where norm denotes activation and each component corresponds to some abstract feature. These elements are generally unordered. Here we use GSR to order these features consistently between digits. We train a capsule net on MNIST with GSR on 16 fully connected graphs between the 10 digit capsules. In the standard capsule network, each capsule orders features randomly based on initialization. However, with GSR we obtain a \textit{consistent feature ordering}, e.g. node 1 corresponds to line thickness across all digits. GSR enforces a more ordered and interpretable encoding where localized regions are similarly organized, and the global line thickness feature is consistently learned between digits. More generally, GSR can be used to order nodes such that features common across capsules appear together. Finally, GSR does not degrade performance much, as can be seen by the digit reconstructions in Fig.~\ref{fig:caps_fig}.

In these examples the goal was to enforce a specified structure on unstructured features, but next we will examine the case where the goal is to learn the structure of the reduced feature space.


\subsection{Learning Graph Structure}
Using the procedure defined in Sec.~\ref{sec:alg:learning}, we can learn a graph structure. We first show that depending on the data, the learned graph exhibits either cluster or trajectory structure. We then show that our framework can learn structures that are hierarchical, i.e. subclusters within clusters or trajectories within clusters. Hierarchies are a difficult structure for other interpretability methods to learn~\cite{esmaeili_structured_2019}. However, our method naturally captures this by allowing for arbitrary graph structure among neurons in a layer.

\begin{figure}[hbt]
    \begin{center}
    \includegraphics[width = 0.8\linewidth]{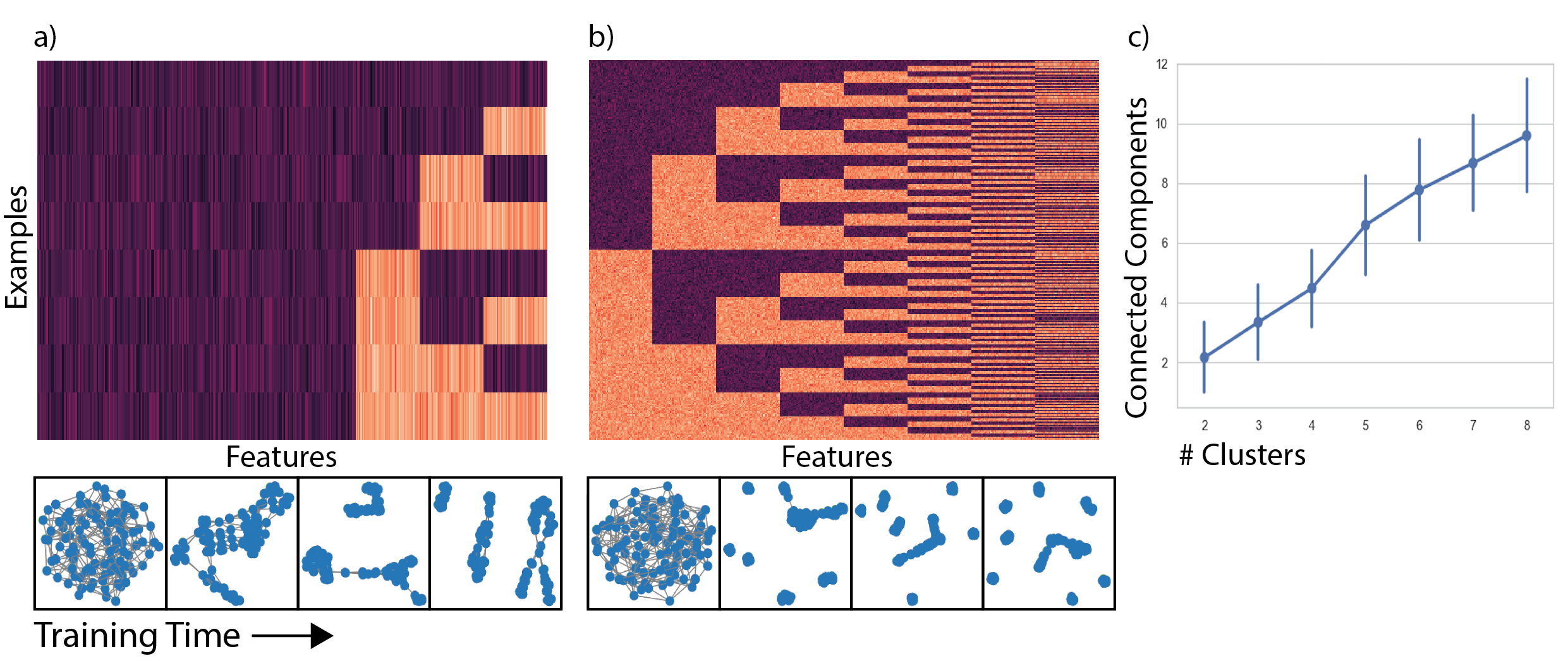}
    \end{center}
    \caption{We show the structure of the training data and snapshots of the learned graph for (a) three modules and (b) eight modules. (c) shows we have the mean and 95\% CI of the number of connected components in the trained graph for over 50 trials.}
    \label{fig:modular}
\end{figure}
\subsubsection{Cluster Structure on Generated Data}~\label{sec:results:cluster}
We structure our $n^{th}$ dataset to have exactly $n$ feature clusters. We generate the data with $n$ clusters by first creating $2^n$ data points representing the binary numbers from $0$ to $2^n -1$, then added gaussian noise $N(0, 0.1)$. This creates a dataset with a ground truth number of feature clusters. In the $n^{th}$ dataset the learned graph should have $n$ connected components for $n$ independent features. In Fig.~\ref{fig:modular} (a-b) we can see how this graph evolves over time for 3 and 8 modules. (c) shows how the learned graph learns the correct number of connected components for each ground truth number of clusters.

\begin{figure}[hbt]
    \begin{center}
    \includegraphics[width = 0.9\linewidth]{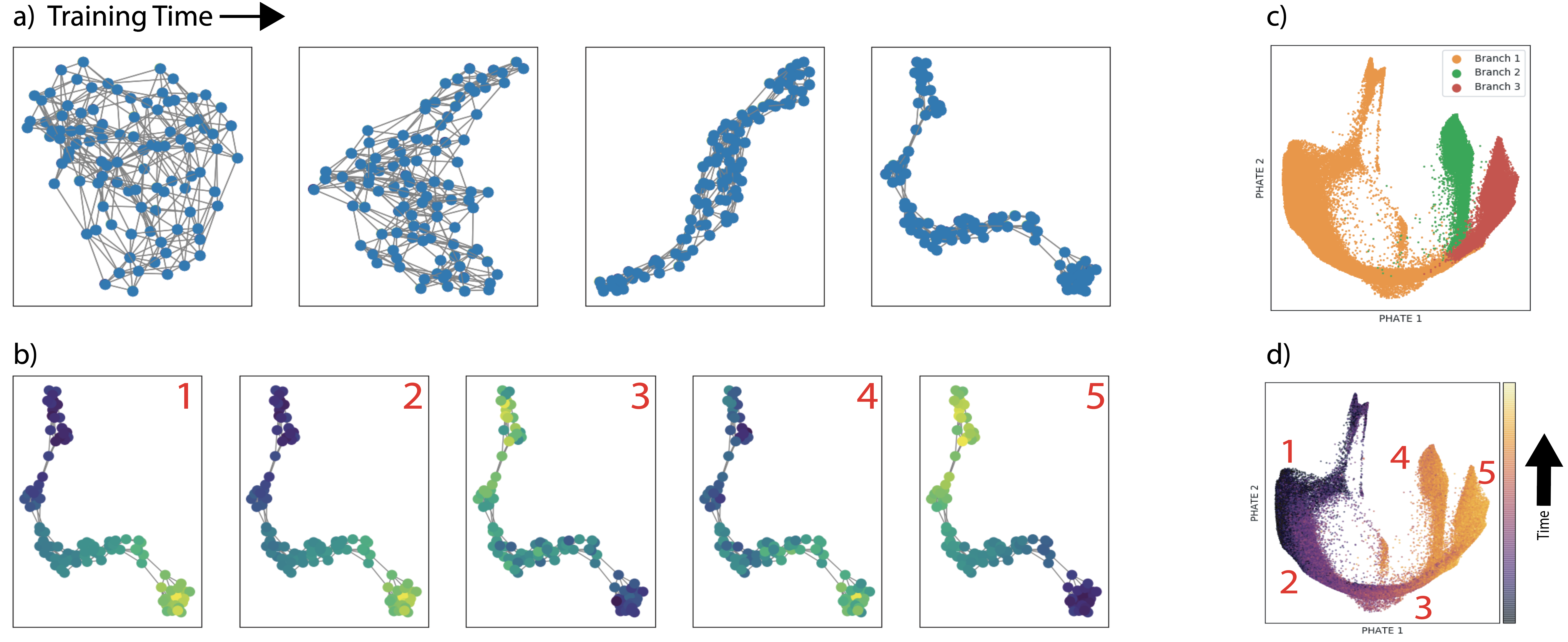}
    \end{center}
    \caption{Shows (a) graph structure over training iterations (b) feature activations of parts of the trajectory. PHATE~\cite{moon2017phate} embedding plots colored by (c) branch number and (b) inferred trajectory location showing the branching structure of the data.}
    \label{fig:wishbone}
\end{figure}
\subsubsection{Trajectory Structure on T cell Development Data.}~\label{sec:results:wishbone}
Next, we test graph learning on biological mass cytometry data, which is a high dimensional, single-cell protein dataset, measured on differentiating T cells from the Thymus~\cite{setty2016wishbone}. The T cells lie along a bifurcating progression where the cells eventually diverge into two lineages (CD4+ and CD8+). Here, the structure of the data is a trajectory (as opposed to a pattern of clusters). We can see in Fig.~\ref{fig:wishbone} how the activated nodes in the graph embedding layer correspond to locations along the data trajectory, and importantly, the learned graph is a single connected component. The activated nodes (yellow) move from the bottom of the embedding to the top as T-cells develop into CD8+ cells. The CD4+ lineage is also CD8- and thus looks like a mixture between the CD8+ branch and the naive T cells. The learned graph structure here has captured the transitioning structure of the underlying data.

\begin{figure}[t]
    \begin{center}
    \includegraphics[width = .9\linewidth]{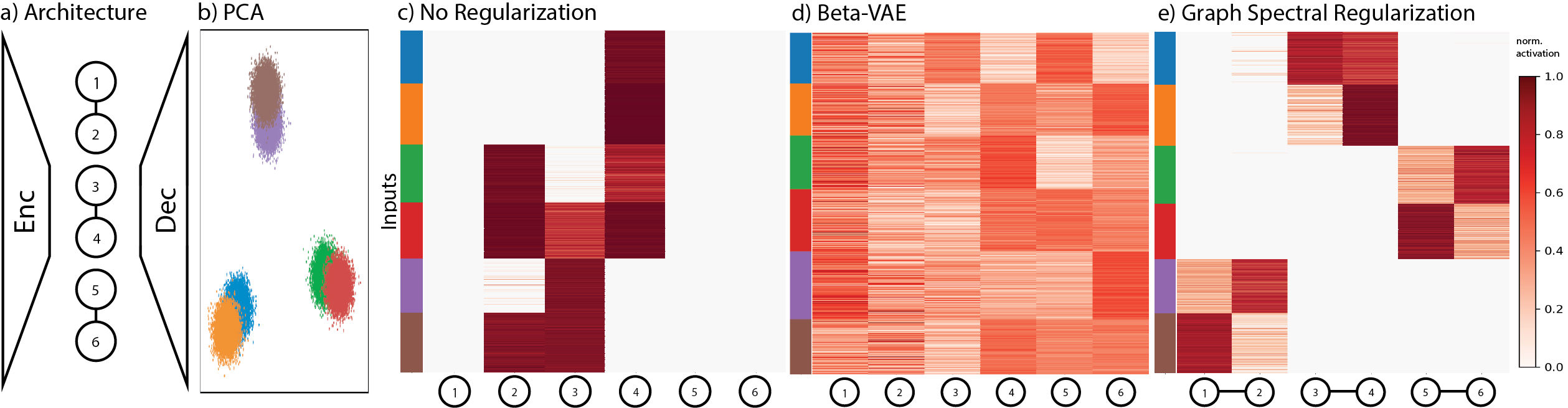}
    \end{center}
    \caption{Graph architecture, PCA plot, activation heatmaps of a standard autoencoder, $\beta$-VAE~\cite{higgins_-vae_2017} and a graph regularized autoencoder. With relu activations normalized to $[0,1]$ for comparison. In the model with graph spectral we are able to clearly decipher the hierarchical structure of the data, whereas with the standard autoencoder or the $\beta$-VAE the structure of the data is not clear.}
    \label{fig:hierarchical}
\end{figure}

\subsubsection{Clusters within Clusters on Generated Data.}
We demonstrate graph spectral regularization on data that is generated with a structure containing sub-clusters. Our data contains three large-scale structures, each comprising two Gaussian sub clusters generated in 15 dimensions (See Fig. \ref{fig:hierarchical}). We use this dataset as it has both global and local structure. We demonstrate that our graph spectral regularized model is able to pick up on both the global and local structure of this dataset where disentangling methods such as $\beta$-VAE cannot. We use a graph-structure layer with six nodes with three connected node pairs and employ the graph spectral regularization. After training, we find that each node pair acts as a ``super node" that detects each large-scale cluster. Within each super node, each of the two nodes encodes one of each of the two Gaussian substructures. Thus, this specific graph topology is able to extract the hierarchical topology of the data.

\begin{figure}[!ht]
    \begin{center}
    \includegraphics[width = 0.9\linewidth]{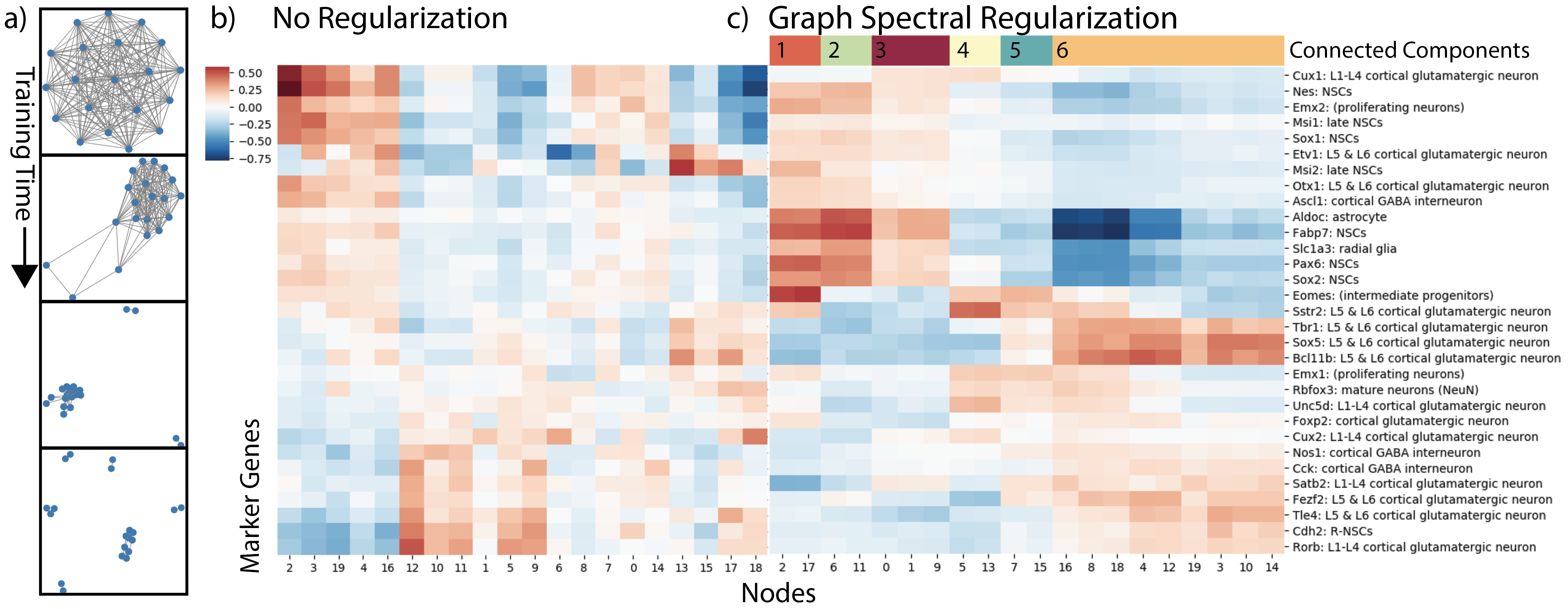}
    \end{center}
    \caption{Shows correlation between a set of marker genes for specific cell types and embedding layer activations. First with the standard autoencoder, then our autoencoder with graph spectral regularization. The left heatmap is biclustered, the right heatmap is grouped by connected components in the learned graph. We can see progression especially in the largest connected component where features on the right of the component correspond to less developed neurons.}
    \label{fig:mouse}
\end{figure}
\subsubsection{Hierarchical Cluster and Trajectory Structure on Developing Mouse Cortex Data.}
In Fig.~\ref{fig:mouse} we learn a graph on a single-cell RNA-sequencing dataset of over 4000 cells and over 8000 genes. The data contains a set of cells in the process of developing from neural stem cells to full neurons in the mouse brain. While there are many gene modules that contribute to the neuronal development, there are some states that have been studied. We use a list of cell type marker genes to validate our method. We use 1000 PCA components of the data in an autoencoder with a 20-dimensional embedding space. We learn the graph using an adaptive bandwidth gaussian kernel with the bandwidth for each feature set to the Euclidean distance to the nearest neighboring feature.

Our graph learns six components that represent meta features over the gene space. We can identify each with a specific type of cell or related types of cells. For example, the light green component (cluster 2) represents the very early stage neural stem cells as it is highly correlated with increased Aldoc, Pax6 and Sox2 gene expression. Most interesting to examine is cluster 6, the largest component, which represents development into mature neurons. Within this component we can see a progression from just after intermediate progenitors on the left (showing Eomes expression) to more mature neurons with higher expression of Tbr1 and Sox5. With a standard autoencoder we cannot see progression structure of this dataset. While some of the more global structure is captured, we fail to see the data progression from intermediate progenitors to mature neurons. Learning a graph allows us to create receptive fields e.g. clusters of neurons that correspond to specific structures within the data, in this case cell types. Within these neighborhoods, we can pick up on the substructure within a single cell type, i.e. their developmental trajectory.

\subsection{Computational Cost}
Our method can be used to increase interpretability without much loss in representation power. At low levels, GSR can be thought of as rearranging the activations so that they become spatially coherent. As with other interpretability methods, GSR is not meant to increase representation power, but create useful representations with low cost in power. Since GSR does not require an information bottleneck such as in $\beta$-VAE, a GSR layer can be very wide, while still being interpretable. In comparing loss of representation power, GSR should be compared to other regularization methods, namely L1 and L2 penalties (See Table~\ref{tab:power}). In all three cases we can see that a higher penalty reduces the model capacity. GSR affects performance in approximately the same way as L1 and L2 regularizations do. To confirm this, we ran a MNIST classifier and measured train and test accuracy with 10 replicates. 
\begin{table}[tbh]
\centering
\begin{tabular}{r | l l l}
Regularization& Training accuracy & Test Accuracy & Coefficient \\ \hline
None & $99.1 \pm 0.3$ & $97.5 \pm 0.3$ & N/A       \\ 
L1   & $98.9 \pm 0.3$ & $97.4 \pm 0.4$ & $10^{-4}$ \\ 
L2   & $98.3 \pm 0.3$ & $98.0 \pm 0.2$ & $10^{-4}$ \\ 
GSR (ours)  & $99.3 \pm 0.3$ & $98.0 \pm 0.3$ & $10^{-3}$ \\
\end{tabular}
\caption{MNIST classification training and test accuracies for coefficient selected using cross validation over regularization weights in $[10^{-7}, 10^{-6}, \ldots, 10^{-2}]$ for various regularization methods with standard deviation over 10 replicates.}
\label{tab:power}
\end{table}
Graph spectral regularization adds a bit more overhead than elementwise activation penalties. However, the added cost can be seen as containing one matrix vector operation per pass. Empirically, GSR shows similar computational cost as other simple regularizations such as L1 and L2. To compare costs, we used a Keras model with Tensorflow backend~\cite{abadi_tensorflow_2016} on a Nvidia Titan X GPU and a dual Intel(R) Xeon(R) CPU E5-2697 v4 @ 2.30GHz, and with batchsize 256. we observed during training 233 milliseconds (ms) per step with no regularization, 266ms for GSR, and 265ms for L2 penalties. 

\section{Conclusion}
\label{sec:conclusion}
We have introduced a novel biologically inspired method for regularizing features of the internal layers of dense neural networks to take the shape of a graph. We show that coherent features emerge and can be used to interpret the underlying structure of the dataset. Furthermore, when the intended graph is not known apriori, we have presented a method for learning the graph structure, which learns a graph relevant to the data. This regularization framework takes a step towards more interpretable neural networks, and has applicability for future work seeking to reveal important structure in real-world biological datasets as we have demonstrated here.

\section*{Acknowledgements}
This research was partially funded by IVADO (l'institut de valorisation des donn\'{e}es) [\emph{G.W.}]; Chan-Zuckerberg Initiative grants 182702 \& CZF2019-002440 [\emph{S.K.}]; and NIH grants R01GM135929 \& R01GM130847 [\emph{G.W.,S.K.}].
\begin{spacing}{0.94}
\bibliography{auto,manual}
\bibliographystyle{splncs04}
\end{spacing}

\end{document}